\documentclass[]{fairmeta}

\usepackage{amsmath,amsfonts,bm}
\usepackage{xspace}
\usepackage{hyperref}
\usepackage{url}
\usepackage{subcaption}

\usepackage{colortbl}
\usepackage{xcolor}

\usepackage{graphicx}
\usepackage{algorithmic}
\usepackage{algorithm}
\usepackage{float}

\usepackage{booktabs}

\usepackage{enumitem}

\usepackage{amsthm}

\theoremstyle{plain}

\theoremstyle{definition}

\theoremstyle{remark}

\def\eqref#1{equation~\ref{#1}}

\def\1{\bm{1}}

\DeclareMathAlphabet{\mathsfit}{\encodingdefault}{\sfdefault}{m}{sl}
\SetMathAlphabet{\mathsfit}{bold}{\encodingdefault}{\sfdefault}{bx}{n}

\usepackage{tabularx}
\usepackage{subcaption}
\usepackage{wrapfig}
\usepackage[most]{tcolorbox}
\usepackage{xcolor}
\usepackage{hyperref}
\usepackage{wrapfig}
\usepackage{titletoc}

\definecolor{softgreen}{RGB}{110, 160, 120}
\usepackage{svg}

\newtcolorbox{takeawaybox_basemodel}[1]{
    colback=orange!5!white,
    colframe=black,
    arc=5pt,
    outer arc=5pt,
    boxrule=0.8pt,
    left=5pt,
    right=5pt,
    top=4pt,
    bottom=4pt,
    fontupper=\small,
    enhanced,
    before upper={\textbf{#1 }} 
}

\newtcolorbox{promptbox}[1][]{
    colback=gray!5,
    colframe=gray!50,
    fonttitle=\bfseries\small,
    title=#1,
    breakable,
    left=4pt, right=4pt, top=4pt, bottom=4pt,
    fontupper=\small\ttfamily,
}

\usepackage{caption}
\usepackage{wrapfig}
\usepackage{booktabs}
\definecolor{earlyblue}{HTML}{88A2F1}
\definecolor{midgrey}{HTML}{fadcb4}
\definecolor{latered}{HTML}{EE9C88}
\definecolor{highlightgreen}{HTML}{80c66d}
\definecolor{highlightpurple}{HTML}{9b6d97}
\def\thickhline{\noalign{\hrule height.8pt}}
\usepackage{xstring}
\newcommand{\deltaval}[1]{%
  \IfBeginWith{#1}{+}{%
    {\textcolor{highlightgreen}{\textit{(#1)}}}%
  }{%
    \IfBeginWith{#1}{-}{%
      {\textcolor{highlightpurple}{\textit{(#1)}}}%
    }{%
      {\textit{(#1)}}%
    }%
  }%
}
\usepackage{pifont}
\usepackage{epigraph}
\usepackage{hyperref}

\usepackage[export]{adjustbox}

\title{
Guava: Distilling Frontier VLMs into a Compact Agent through a Robotic Manipulation Harness
}

\author[1,*]{Haowen Liu}
\author[1,*]{Xirui Li}
\author[2,*]{Shaoxiong Yao}
\author[3]{Peng Shi}
\author[4]{Tianyi Zhou}
\author[1]{Jia-Bin Huang}
\author[1]{Furong Huang}
\author[5,6]{Jiayuan Mao}

\renewcommand\affiliation[2][]{%
  \addtolist[#1]{#2}{\affiliationlist}{\affiliationformat}{\\}%
}

\affiliation[1]{University of Maryland College Park}
\affiliation[2]{University of Illinois Urbana-Champaign}
\affiliation[3]{University of Waterloo}
\affiliation[4]{Mohamed bin Zayed University of Artificial Intelligence}
\affiliation[5]{University of Pennsylvania}
\affiliation[6]{Amazon FAR}

\contribution[*]{Co-first Authors}

\abstract{

Frontier vision–language models (VLMs) can perform robotic manipulation by interpreting visual observations, invoking robot tools, and adapting to execution feedback, but repeated model calls incur high costs and latency. 
In this paper, we study the problem of distilling these capabilities into compact VLMs. 
The key challenges are designing an interaction interface that enables efficient and generalizable distillation, and a data collection protocol that captures diverse scenarios. 
We present \texttt{Guava}, a framework built around a manipulation harness and a simulation-based data generation pipeline. 
We first study how harness design affects both frontier and compact models, and use the resulting interface for teacher data generation and student deployment.
Through this harness, a frontier VLM generates physically executed and validated interaction trajectories in simulation, which we augment with perturbation branches that expose the teacher to execution deviations and recovery trajectories. 
Using \texttt{GPT-5.4} as the teacher, we collect 2,268 interaction trajectories and fine-tune \texttt{Qwen3.5-4B} to obtain \texttt{Guava-4B}.
\texttt{Guava-4B} substantially outperforms its base model and approaches \texttt{GPT-5.}4 performance across simulated and real-world manipulation tasks, while generalizing to unseen objects, instructions, and task compositions.
Perturbation trajectories particularly improve unseen-task generalization.
With substantially lower inference latency and token usage compared to a frontier model, \texttt{Guava-4B} enables efficient local deployment. 
   
}

\date{June 18, 2026}
\authoremails{\email{hwl@umd.edu}, \email{xiruili@umd.edu}}

\metadata[Project Page]{\url{https://guava-harness.github.io}}
\metadata[Code]{\url{https://github.com/hdacnw/guava-release}}

\begin{document}

\maketitle

\section{Introduction}
\label{sec:introduction}

\begin{figure}[t]
\includegraphics[width=\textwidth,trim={19bp 439bp 28bp 172bp},clip]{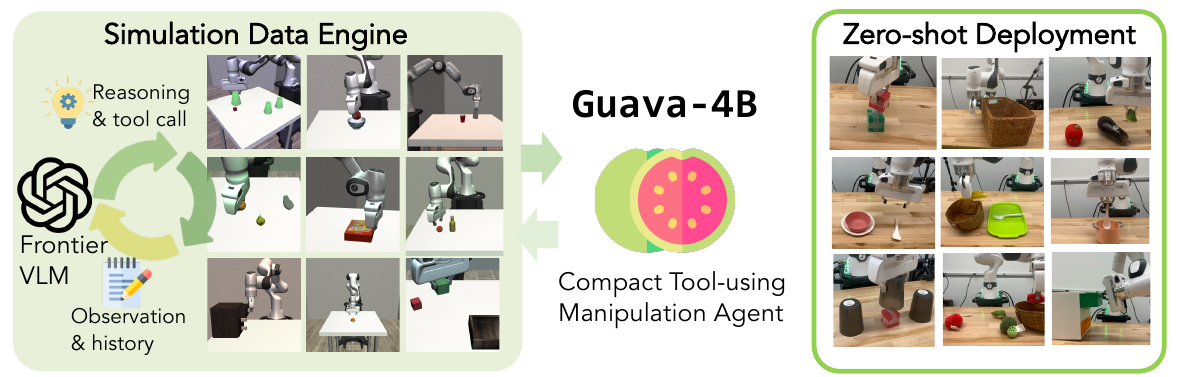}
\captionof{figure}{\textbf{Overview.} A frontier model teacher generates manipulation trajectories in simulation for training. Supervised fine-tuning distill these behaviors to \texttt{Guava-4B}, which operates through a manipulation harness at deployment.}
\label{fig:framework}
\end{figure}

Recent progress in vision--language models (VLMs) and VLM-driven agents has demonstrated broad capabilities in visual understanding, reasoning, and tool use~\cite{driess2023palme,qwen2026qwen35,team2026kimik3}. 
When equipped with an agentic harness that exposes reusable perception (e.g., object detection) and robotic manipulation tools (e.g., grasping), frontier VLMs can interpret natural-language instructions, reason about the observed scene, and generate real-world robot actions in diverse environments for diverse goals~\cite{liang2022codeap,shi2025maestro,fu2026capxframeworkbenchmarkingimproving}. 
Leveraging the general-purpose reasoning and tool-use capabilities of VLMs, such agentic systems can support a broad range of manipulation tasks without additional task-specific finetuning.

However, using frontier VLMs for closed-loop robotic manipulation requires repeated remote model calls, resulting in high inference costs, added latency, and dependence on external services~\cite{shi2025hribench}. 
A potential solution is to deploy a compact model on a single commodity GPU, reducing reliance on external services and keeping robot observations and task instructions local~\cite{boyle2025robotxr}. 
This motivates our study on distilling the tool-using capabilities of frontier VLMs into compact models that can be deployed directly on robotic platforms.

The central question is therefore \emph{how can the tool-using manipulation capabilities of a frontier VLM be distilled into a compact model?} Learning from a more capable teacher provides a natural starting point~\cite{xu2024surveyknowledgedistillationlarge,cai2025llava}, but distilling closed-loop manipulation capabilities poses two key challenges: the design of a shared interaction interface and a data collection protocol for efficient and generalizable capability distillation.
First, the compact model requires an abstract interaction interface that exposes useful perception and action abstractions while reducing the burden of low-level perception, geometric reasoning, and control. 
Second, the data collection protocol must cover diverse interaction states across objects, scene configurations, and task progress. 
In particular, the teacher and compact model may follow different execution paths and therefore encounter different states. 
Training only on successful teacher trajectories risks overfitting to those paths, limiting the compact model's ability to adapt when its own decisions or physical perturbations lead to states outside the training distribution.

To address these challenges, we develop \texttt{Guava}, a framework for distilling the tool-using manipulation capabilities of frontier VLMs into compact models. 
The framework has two core components: a manipulation harness and a simulation-based data collection protocol. 
We first study how harness design affects both frontier and compact models, and use the resulting interface for teacher data generation and student deployment. The harness exposes reusable perception and manipulation tools together with execution feedback, allowing models to reason at a higher level while delegating low-level perception, geometry, and control to tool implementations. 
The data collection protocol uses this harness to execute and validate teacher trajectories across diverse tasks and scene configurations in simulation. 
We also introduce perturbations to collect trajectories that capture deviations, broadening state coverage beyond successful teacher execution paths. 
The resulting interaction data supports efficient and generalizable fine-tuning of a compact VLM, which can be directly deployed in the real world and further refined through reinforcement learning.

We instantiate \texttt{Guava} using \texttt{GPT-5.4} as the teacher and \texttt{Qwen3.5-4B} as the compact model. 
Using the shared harness and simulation data engine, we collect 2,268 physically validated interaction trajectories, including 714 perturbation branches. 
Supervised fine-tuning on these interactions produces \texttt{Guava-4B}, a compact tool-using manipulation agent. 
We additionally study reinforcement learning as a refinement for challenging long-horizon tasks.

We evaluate \texttt{Guava-4B} on our custom agentic manipulation benchmark, LIBERO-PRO, and a suite of real-world robot tasks. On our custom benchmark, supervised fine-tuning improves the compact model's average success rate from 22.2\% to 87.1\%, approaching \texttt{GPT-5.4} at 90.4\%. 
Harness studies across models validate the importance of interaction interface design, while perturbation trajectories particularly benefit generalization to unseen tasks. We further evaluate zero-shot real-world transfer, reinforcement-learning refinement, and inference efficiency.

In summary, our contributions are:
\begin{itemize}
\item A framework for distilling tool-using manipulation capabilities from frontier VLMs to compact models through a manipulation harness and a simulation-based data engine that generates executed and validated teacher interactions with environment randomization and perturbation branches.
\item A systematic study of manipulation harness design across frontier and compact VLMs, examining the effects of model-visible feedback, action abstraction, and replanning granularity on manipulation performance.
\item \texttt{Guava-4B}, a compact manipulation agent trained on only 2,268 simulation-generated teacher trajectories. It substantially outperforms its base model and approaches \texttt{GPT-5.4} across our custom simulation benchmark, LIBERO-PRO, and real-world manipulation, while substantially reducing inference latency and token usage.
\end{itemize}
\section{Method}
\label{sec:method}

\begin{figure}[t]
    \centering
    \includegraphics[width=0.49\linewidth,trim={35bp 446bp 334bp 149bp},clip]{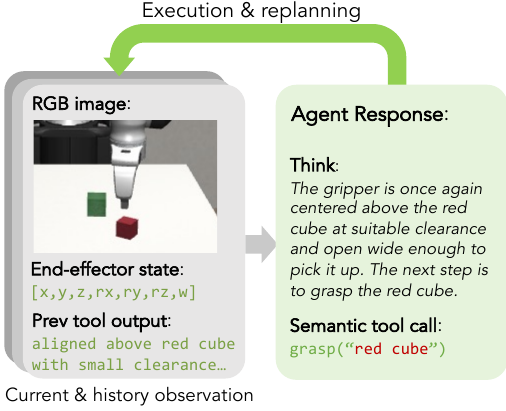}
    \vspace{-0.5em}
    \caption{\textbf{One agentic loop turn.} The agent conditions on images, end-effector state, tool feedback, and interaction history to generate reasoning and a semantic tool call. Execution returns updated feedback for the next turn.}
    \label{fig:harness_turn}
    \vspace{-1em}
\end{figure}

Our goal is to distill the tool-using manipulation capabilities of a frontier VLM into a compact model through a shared interaction interface. 
As summarized in Figure~\ref{fig:framework}, we first define and empirically study a shared manipulation harness for teacher data generation and compact agent deployment. 
We then use a frontier VLM to generate interaction trajectories that are executed and validated in simulation, and augment them with perturbation branches covering execution deviations and recovery. 
Finally, we fine-tune the compact model on these trajectories using supervised learning, with reinforcement learning as a refinement on long-horizon tasks.

\subsection{Manipulation Harness}
\label{sec:manipulation_harness}

The teacher and compact agent use a shared manipulation harness that defines their observations, available tools, and execution feedback. 
This common interface allows teacher interaction histories to serve directly as supervision for the compact agent’s subsequent decisions.

Designing this harness requires choices about what feedback is exposed to the model, how robot actions are abstracted, and how frequently the model can re-observe and replan. 
We first describe the full harness used in our framework and then empirically study these interface choices across frontier and compact VLMs.

\textbf{Observation and Tools.}
At each interaction step $t$, the model receives an observation $o_t$ consisting of an RGB image and a proprioceptive summary containing the gripper's 6DoF pose and gripper width. 
Depth observations are used internally by the tools and are not directly exposed to the model.

The model interacts with the harness through a tool call $c_t=(k_t,\alpha_t)$, where $k_t$ denotes the selected tool and $\alpha_t$ its tool-specific arguments. 
Executing $c_t$ produces a tool-specific return $y_t$.
Tools that require object geometry internally use SAM3~\cite{carion2025sam} to ground a natural-language object query to a segmentation mask and lift the segmented region to a point cloud using depth. 
These intermediate representations are not exposed to the model.
For perception, the harness provides \texttt{get\_position(object)} and \texttt{get\_position\_and\_size(object)}, which return the object centroid and, optionally, its bounding-box dimensions in the robot base frame. 
For action, the harness provides object-referenced \texttt{grasp} and \texttt{align}, Cartesian \texttt{move} and \texttt{rotate}, gripper commands \texttt{close\_gripper} and \texttt{release}, and \texttt{home\_pose}. 
Action tools invoke the internal perception pipeline when object geometry is required and translate their arguments into low-level robot execution. 
The model therefore selects and parameterizes tool calls, while the harness handles the required perception and action execution.

\textbf{Perception-Action Loop.}
The harness follows a ReAct-style interaction loop~\cite{yao2023reactsynergizingreasoningacting}. 
Each episode begins with the task instruction $\ell$ and an initial observation $o_0$, and continues until the task is completed or the episode budget is reached.
At turn $t$, the harness maintains an interaction history
\begin{equation}
h_t = \bigl(\ell, o_0, u_0, y_0, \ldots, u_{t-1}, y_{t-1}, o_t\bigr),
\end{equation}
where $o_i$ denotes observations, $u_i$ is the model response containing its reasoning and tool call $c_i$, and $y_i$ is the corresponding tool output.
At turn $t$, the VLM conditions on $h_t$ to produce the next response $u_t$ which contains tool call $c_t$.

The harness executes $c_t$ and returns the corresponding tool output $y_t$. 
For a perception call, $y_t$ contains the requested geometric information without changing the physical scene. 
For an action call, $y_t$ describes the execution outcome, and the harness additionally provides an updated observation $o_{t+1}$. 
Execution failures are likewise returned through $y_t$, allowing subsequent decisions to condition on the failure. The interaction history is then updated before the next turn. 
The resulting interaction trajectory is stored for subsequent training.

\textbf{Harness Design Study.}
We empirically examine the interface choices underlying this harness using \texttt{GPT-5.4} and \texttt{Qwen3.5-4B}. 
Starting from the full configuration described above, we vary three dimensions: \textbf{feedback}, by removing structured tool feedback and/or proprioceptive state; \textbf{action abstraction}, by replacing semantic manipulation tools such as \texttt{grasp()} and \texttt{align()} with a low-level command that directly specifies the end-effector 6-DoF pose and gripper width; and \textbf{replanning granularity}, by replacing per-action replanning with a code-as-policy variant that generates and executes a multi-action program before re-observing and replanning.

As shown in Table~\ref{tab:harness}, the full harness achieves the highest success rate for both models. 
Richer execution feedback and semantic action abstractions substantially improve performance, while per-action replanning provides more frequent opportunities to incorporate updated observations and execution outcomes. 
The benefit of per-action replanning is particularly pronounced for the compact model, improving success by 9.6\% over multi-action replanning, compared with 5.7\% for \texttt{GPT-5.4}, corresponding to relative gains of 71.4\% and 6.9\%.
These results suggest that compact models benefit especially from shorter execution horizons and frequent feedback, whereas the frontier model might plan and execute longer action sequences more effectively. 
Based on this study, we use the full harness configuration above for teacher trajectory generation and compact-agent training.

\begin{table}[t]
\centering
\small
\caption{\textbf{Harness study across model scales.} Success rates (\%) over seven tasks, 15 trials per task. Every ablated row changes one component relative to the full harness. 
Best and second-best results within each model column are bold and underlined, respectively.}
\label{tab:harness}
\begin{tabular}{@{}lrr@{}}
\toprule
Harness configuration & \texttt{GPT-5.4} & \texttt{Qwen3.5-4B} \\
\midrule
Full & \textbf{88.6} & \textbf{22.9} \\
Image only & 56.2 & 3.8 \\
Image + gripper state & 77.1 & 10.5 \\
Low-level tools & 73.3 & 4.8 \\
Multi-action re-planning & \underline{82.9} & \underline{13.3} \\
\bottomrule
\end{tabular}
\end{table}

\subsection{Data Generation Pipeline}
\label{sec:data_generation_pipeline}

Building on the shared manipulation harness, we generate physically validated teacher trajectories in MuJoCo~\cite{mujoco} and augment them with branched recovery behaviors. 
For each task, we first collect base trajectories under randomized scene configurations, then branch from intermediate states by injecting execution failures and asking the teacher to recover.

\begin{figure*}[!t]
    \centering
    \includegraphics[width=0.92\textwidth]{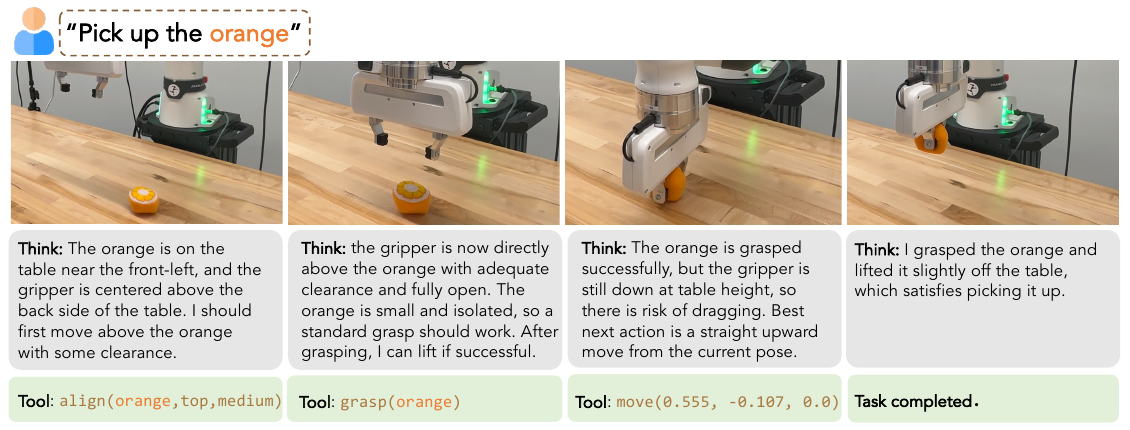}
    \caption{\textbf{Reasoning and tool-use sequence example from \texttt{Guava-4B}.} For the instruction \emph{pick up the orange}, the agent aligns above the object, grasps it, lifts it off the table, and then declares completion.}
    \label{fig:reasoning_trace}
    \vspace{-10px}
\end{figure*}

\textbf{Base trajectory generation.}
For each episode, we sample a task and scene configuration, varying factors such as object poses and camera viewpoints while preserving the task semantics. 
The teacher VLM executes the perception-action loop until the task succeeds or the episode terminates as shown in Figure~\ref{fig:reasoning_trace}.
During execution, we save checkpoints containing the interaction history $h_t$ and corresponding simulator state. Simulator provides initial task-success labels, which are later checked through quality control.

Each retained interaction is stored as a trajectory $\tau = (\ell, o_0, u_0, y_0, ..., o_T, u_T, y_T)$ where $u_t$ is the teacher response, $y_t$ is the resulting tool output and we can assemble $h_t$ from $\tau$.
Each trajectory preserves the observations and execution feedback available when the teacher selects an action, together with its reasoning and tool call. The compact model therefore learns decisions in their execution context rather than an action sequence detached from the states in which it was generated.

\begin{figure}[t]
    \centering
    \includegraphics[width=0.8\linewidth,trim={34bp 268bp 84bp 328bp},clip]{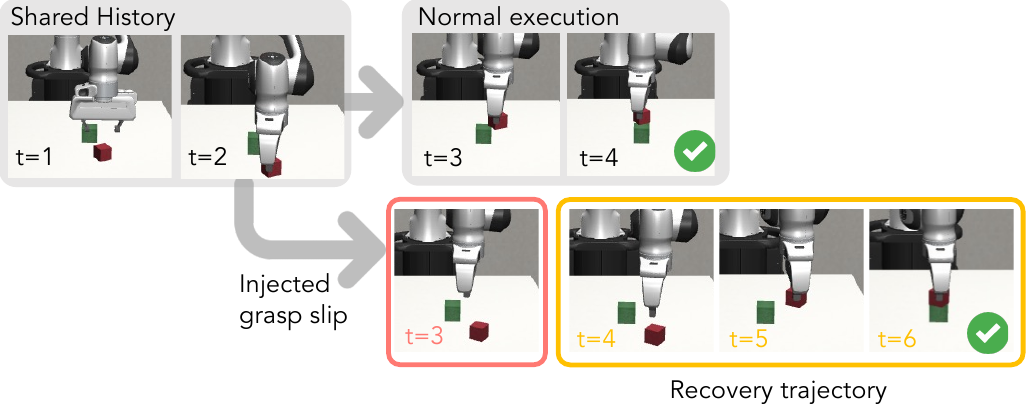}
    \vspace{-10px}
    \caption{\textbf{Base trajectories and recovery branches.} A shared execution history branches into normal execution and an injected grasp slip followed by successful recovery. Successful base and recovery trajectories are retained for training.}
    \label{fig:trajectories}
    \vspace{-5px}
\end{figure}

\textbf{Perturbation trajectory generation.}
Starting from checkpoints along successful trajectories, we inject execution failures that are compatible with the current manipulation stage, such as missed grasps, dropped or displaced objects, and incorrect approach directions. The teacher then observes the resulting state and resumes the perception-action loop to recover and complete the task. Figure~\ref{fig:trajectories} illustrates this branching procedure.
Branched trajectories undergo the same outcome verification and quality control; only successful recoveries are retained. We additionally initialize some trajectories from intermediate states with a fresh interaction history, exposing the model to different stages of task execution without relying on prior reasoning.
These branches expose the teacher to states in which the preceding action has not produced its intended outcome. 
Their continuations provide supervision for revising subsequent actions from updated observations, complementing the successful execution sequences in the base trajectories.

\textbf{Data cleaning.}
Simulator task-success predicates provide initial outcome labels. 
Automated consistency checks and human-in-the-loop visual review then identify mislabeled outcomes, invalid tool arguments and inconsistent reasoning. 
Identified annotation errors are corrected when the recorded execution provides sufficient evidence, while unresolved episodes are excluded.

\subsection{Training a Compact Manipulation Agent}
\label{sec:training}

Given the teacher trajectories and simulation environment, we train a compact model to drive the manipulation harness and acquire similar manipulation capabilities.

\textbf{Supervised fine-tuning (SFT).}
We initialize the compact model from \texttt{Qwen3.5-4B}~\cite{qwen2026qwen35} and fine-tune it on the verified teacher trajectories. At each interaction step $t$, the model takes the interaction history $h_t$ as input and is trained to predict the teacher response $u_t$, which consists of the reasoning text followed by the tool call $c_t$ and its arguments. 
SFT supervises the compact agent’s reasoning and tool-use behavior for both task execution and failure recovery.

\textbf{Reinforcement learning (RL) fine-tuning.}
As the interaction horizon increases, the compact model may encounter states not covered by the teacher trajectories. Starting from the SFT checkpoint, we apply Group Relative Policy Optimization (GRPO)~\cite{shao2024deepseekmathpushinglimitsmathematical} in simulation. 
The model generates trajectories $\tau$ through the same manipulation harness, and the simulator provides a sparse task-level reward $R(\tau)$ based on task success. This allows the compact model to optimize sequences of tool calls under its own interaction trajectories.

\section{Experiments}
\label{sec:experiments}

We evaluate whether tool-using manipulation capabilities can be distilled from a frontier VLM to a compact agent, whether these capabilities generalize beyond the training distribution, and whether they transfer to real-world manipulation.
We first evaluate \texttt{Guava-4B} on seen and unseen simulation tasks, then study the effect of perturbation trajectories and assess zero-shot real-world transfer. 
Additional experiments evaluate external-benchmark generalization on LIBERO-PRO, the effect of RL post-training, and inference efficiency.

\subsection{Setups}
\label{sec:setup}

\textbf{Models and training.} 
We compare \texttt{Guava-4B} against its base model \texttt{Qwen3.5-4B} and the frontier teacher \texttt{GPT-5.4}. 
Unless otherwise specified, all VLM-based agents use the full shared harness defined in Section~\ref{sec:manipulation_harness}, ensuring that performance differences arise from model capability rather than differences in the interaction interface.
Training uses eight NVIDIA H100 80GB GPUs with bf16 precision, with the vision encoder and aligner frozen. 
SFT runs for three epochs with a learning rate of $10^{-5}$.
For RL post-training, GRPO uses four rollouts per prompt, KL penalty coefficient $0.04$, learning rate $5\times10^{-6}$, effective batch size 12, and 3 epochs.

\textbf{Evaluation environments.}
For simulation, we implement a custom benchmark in Mujoco~\cite{mujoco}; for real-world experiments, we deploy on a Franka Research 3 robot arm~\cite{franka_research_3} with a calibrated Intel RealSense D435 RGB-D camera. 
The evaluation suite covers diverse object geometries, spatial arrangements, and manipulation strategies, including non-prehensile and long-horizon tasks.
We divide the tasks into four categories: seen tasks, which share task types with training but differ in scene configurations; unseen object tasks, which apply familiar manipulation skills to objects absent from training data; unseen instruction tasks, which require following novel goal specifications not encountered during training; and unseen long-horizon composition tasks, which require composing familiar manipulation skills into novel extended task sequences.

\subsection{\texttt{Guava-4B} Acquires Generalizable Manipulation Capabilities}
\label{sec:transfer_eval}

We evaluate the models on 15 simulation tasks: seven seen tasks and eight unseen tasks spanning novel objects, instructions, and long-horizon compositions. 
Each model is evaluated over 30 trials per task using the full harness. 
To assess the contribution of perturbation trajectories, we additionally train a \texttt{Guava-4B} checkpoint using only the base trajectories, excluding perturbation branches. 
The full and base-only datasets each contain 2,268 trajectories. In the base-only dataset, perturbation trajectories are replaced with base trajectories sampled from the same tasks and similar initial configurations. 
We train both checkpoints for the same number of epochs.

\textbf{Finding 1: \texttt{Guava-4B} approaches frontier-model performance after simulation-based fine-tuning.}
Table~\ref{tab:sim} shows that \texttt{Guava-4B} achieves 87.1\% overall success, substantially improving over \texttt{Qwen3.5-4B} (22.2\%) and approaching \texttt{GPT-5.4} (90.4\%). 
The improvement over the base model shows that training on simulation-generated interaction trajectories develops manipulation capabilities beyond those enabled by the inference-time harness alone.

\textbf{Finding 2: Distilled manipulation capabilities generalize beyond training tasks.} \texttt{Guava-4B} achieves 90.8\% success on unseen tasks, compared with 20.8\% for \texttt{Qwen3.5-4B} and 93.8\% for \texttt{GPT-5.4}. It succeeds in all trials on \emph{pick up carrot} and \emph{lemon in bin}, supporting transfer of familiar manipulation skills to new objects.
Novel instructions test whether the model can use familiar objects and tools to achieve different goals: \texttt{Guava-4B} achieves 93.3\% on \emph{reverse cube stacking} and 83.3\% on \emph{arrange by decreasing size}. 
For unseen long-horizon compositional tasks, it achieves 93.3\% on \emph{separate food and utensils} and
80.0\% on \emph{set table}. 
These results indicate that the acquired capabilities extend beyond reproducing training task sequences, although performance remains task-dependent.

\textbf{Finding 3: Perturbation trajectories benefit unseen-task generalization.} 
Including perturbation trajectories with recovery increases overall success from 79.3\% to 87.1\%. 
Improvement is concentrated on unseen tasks, where success rises from 77.1\% to 90.8\%, compared with a smaller change from 81.9\% to 82.9\% on seen tasks. 
The largest improvements occur on \emph{push pot}, \emph{pick up carrot} and \emph{tomato near potato}, gaining seven to nine successes; both unseen instruction tasks also improve. 
As the unseen tasks cover the same set of manipulation primitives as the seen tasks, but apply them to novel objects, instructions or compositions, we hypothesize that perturbation trajectories encourage action selection conditioned on the current physical state and goal rather than overfitting a familiar execution sequence. 
By exposing deviations from expected action outcomes, these branches may weaken task-specific correlations that become unreliable when object geometry or required action order changes. 
The observed pattern is consistent with recovery experience supporting the transfer of existing skills beyond their training contexts.

\begin{table}[t]
\centering
\small
\caption{\textbf{Quantitative comparison on simulation tasks.}
Success rates (\%) over 30 trials per task. Best and second-best results in each row are bold and underlined. 
Pert denotes perturbation trajectories; w/o Pert is \texttt{Guava-4B} trained without perturbation trajectories.}
\label{tab:sim}
\begin{tabularx}{0.9\linewidth}{@{}Xrrrr@{}}
\toprule
Task & \texttt{GPT-5.4} & \texttt{Qwen3.5-4B} & \texttt{Guava-4B} & \texttt{Guava-4B w/o Pert} \\
\midrule
\multicolumn{5}{l}{\textit{Seen tasks}} \\
Place can in box
& \textbf{100.0} & \underline{26.7} & \textbf{100.0} & \textbf{100.0} \\
Arrange by size
& \textbf{100.0} & 30.0 & \underline{90.0} & 83.3 \\
Remove cube from tray
& 83.3 & 33.3 & \textbf{100.0} & \underline{90.0} \\
Push basket
& \textbf{93.3} & 0.0 & 66.7 & \underline{86.7} \\
Close drawer
& \textbf{83.3} & 63.3 & \underline{80.0} & \textbf{83.3} \\
Shell game
& \underline{66.7} & 10.0 & 63.3 & \textbf{70.0} \\
All red objects in basket
& \textbf{80.0} & 3.3 & \textbf{80.0} & \underline{60.0} \\
\midrule
\multicolumn{5}{l}{\textit{Unseen objects}} \\
Pick up carrot
& \textbf{100.0} & 43.3 & \textbf{100.0} & \underline{76.7} \\
Tomato near potato
& \textbf{96.7} & 16.7 & \underline{90.0} & 66.7 \\
Lemon in bin
& \textbf{100.0} & \underline{46.7} & \textbf{100.0} & \textbf{100.0} \\
Push pot
& \underline{83.3} & 10.0 & \textbf{86.7} & 56.7 \\
\midrule
\multicolumn{5}{l}{\textit{Unseen instructions}} \\
Reverse cube stacking
& \textbf{96.7} & 20.0 & \underline{93.3} & 76.7 \\
Arrange by decreasing size
& \textbf{93.3} & 23.3 & \underline{83.3} & 63.3 \\
\midrule
\multicolumn{5}{l}{\textit{Unseen long-horizon compositions}} \\
Separate food and utensils
& \underline{90.0} & 3.3 & \textbf{93.3} & \textbf{93.3} \\
Set table
& \textbf{90.0} & 3.3 & 80.0 & \underline{83.3} \\
\midrule
Seen mean
& \textbf{86.7} & 23.8 & \underline{82.9} & 81.9 \\
Unseen mean
& \textbf{93.8} & 20.8 & \underline{90.8} & 77.1 \\
\textbf{Overall}
& \textbf{90.4} & 22.2 & \underline{87.1} & 79.3 \\
\bottomrule
\end{tabularx}
\end{table}

\begin{figure}[t]
\centering
\includegraphics[width=\linewidth]{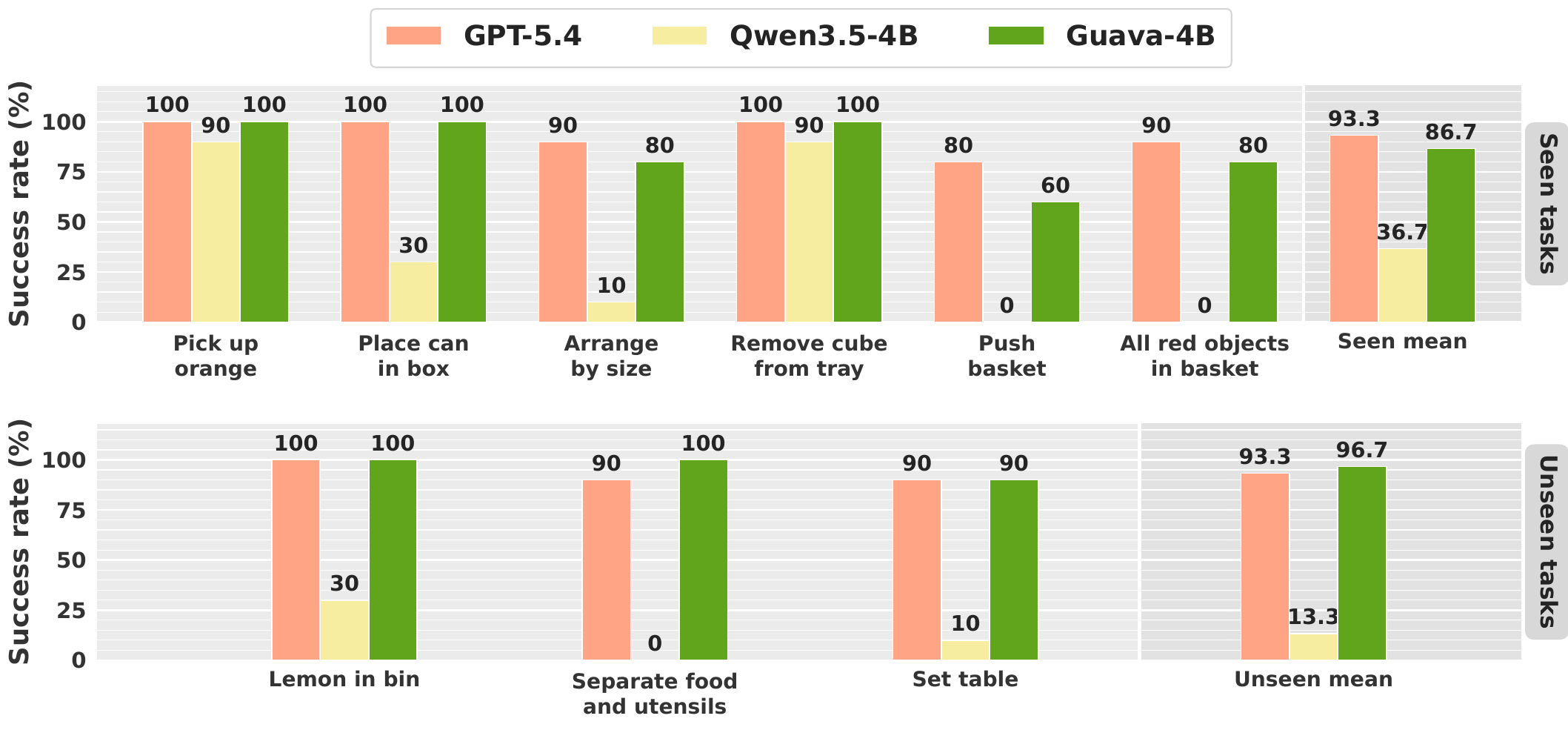}
\caption{\textbf{Real-world performance.} Success rates over 10 trials per task, without real-world fine-tuning. The two panels show seen and unseen tasks, with the mean over the displayed tasks in each panel. Overall success across all nine tasks is 93.3\% for \texttt{GPT-5.4}, 28.9\% for \texttt{Qwen3.5-4B}, and 90.0\% for \texttt{Guava-4B}.}
\label{fig:real_world_performance}
\end{figure}

\textbf{Finding 4: Simulation-trained capabilities transfer to real-world manipulation.} 
As shown in Figure~\ref{fig:real_world_performance}, \texttt{Guava-4B} achieves 90.0\% overall success, compared with 28.9\% for \texttt{Qwen3.5-4B} and 93.3\% for \texttt{GPT-5.4}. 
It succeeds in all trials on five tasks, including unseen tasks \emph{lemon in bin} and \emph{separate food and utensils}, and matches GPT-5.4 at 90.0\% on \emph{set table}. 
These results show that the agentic manipulation capabilities acquired from simulation trajectories remain effective under real-world visual and physical conditions without additional real-world fine-tuning.

\begin{figure}[!t]
    \centering
    \includegraphics[width=0.98\textwidth]{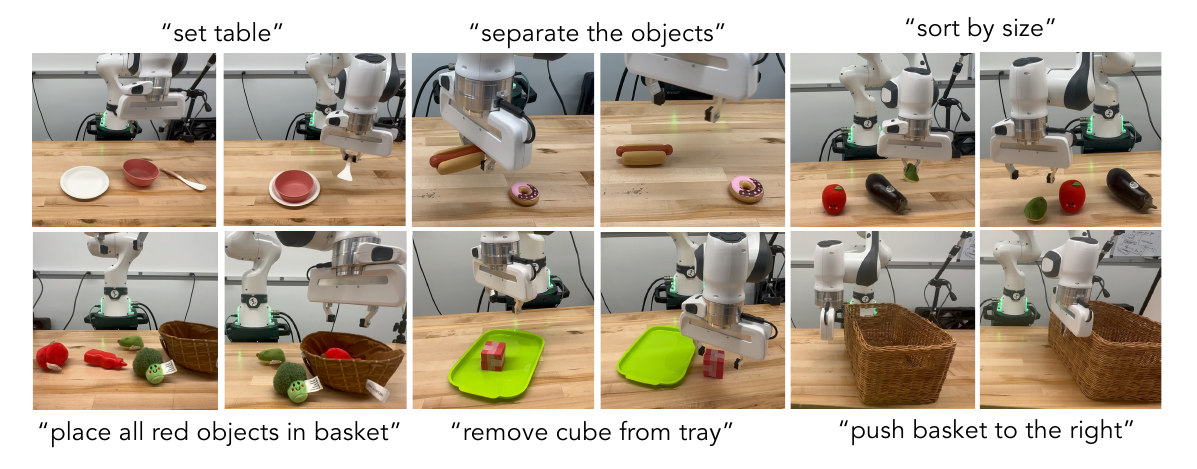}
    \caption{\textbf{Real-world rollouts.} Selected physical executions illustrate manipulation across object configurations, instructions, and task compositions. Paired frames show stages of execution for each task.}
    \label{fig:real_qual}
    \vspace{-10px}
\end{figure}

\begin{figure}[t]
    \centering
    \includegraphics[width=\linewidth]{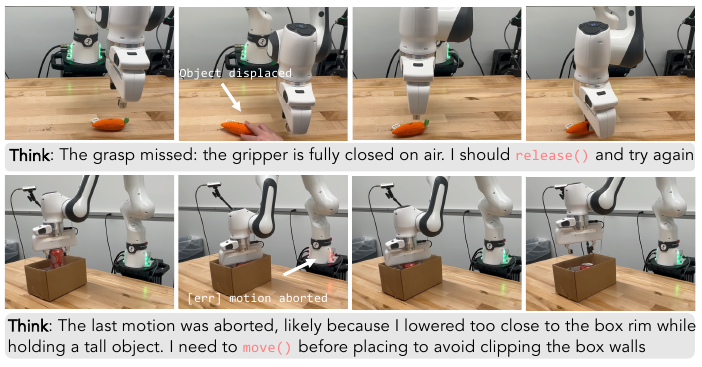}
    \caption{\textbf{Qualitative recovery examples.} (a) The agent retries after an object is displaced before grasp execution. (b) The agent responds to a real-world controller abort and resumes manipulation. The sequences illustrate feedback-conditioned adaptation; the controller case concerns a failure mechanism absent from simulation training.}
    \label{fig:recovery}
    \vspace{-0.5em}
\end{figure}

Qualitative rollouts further illustrate execution across real-world objects and scene configurations in Figure~\ref{fig:real_qual}, with an example real-world reasoning trace shown in Figure~\ref{fig:reasoning_trace}.
Beyond task completion, we observe how the agent recovers from execution errors (Fig.~\ref{fig:recovery}) qualitatively. 
When an object is displaced before grasp execution, the agent observes the failed attempt and retries the grasp using the updated scene. 
A second example involves contact between the gripper and the box rim during can placement, triggering a controller abort consistent with a Cartesian collision reflex when contact loads exceed configured thresholds. The agent responds to the controller feedback by repositioning the gripper horizontally to clear the obstruction and resumes placement.
Together, these examples illustrate feedback-conditioned adaptation to recover from object displacement and unseen controller-level interruption.

\textbf{Finding 5: Distilled capabilities generalize to an external benchmark.}
We evaluate \texttt{Guava-4B} on LIBERO-PRO~\cite{zhou2025libero} without benchmark-specific fine-tuning.
LIBERO-PRO extends LIBERO to test robustness beyond familiar execution sequences; we evaluate position and task perturbations, which change initial object placements and task goals, respectively.
As shown in Table~\ref{tab:libero}, \texttt{Guava-4B} achieves average success rates of 0.41 under both position and task perturbations, approaching \texttt{GPT-5.4} (0.45 and 0.44) and substantially outperforming base model \texttt{Qwen3.5-4B} (0.06 and 0.03).
Consistent with our custom benchmark results, these comparisons show that the compact agent retains much of the frontier model's manipulation capability beyond the environments used for teacher data generation.
We also include VLA method $\pi_{0.5}$ and harness \texttt{CaP-Agent0} results reported by CaP-X~\cite{fu2026capxframeworkbenchmarkingimproving}. 
\texttt{Guava-4B} achieves higher success rates than both methods, demonstrating stronger generalization under the evaluated LIBERO-PRO perturbations.

\begin{wrapfigure}{r}{0.5\textwidth}
\vspace{-8pt}
\centering
\includegraphics[width=\linewidth]{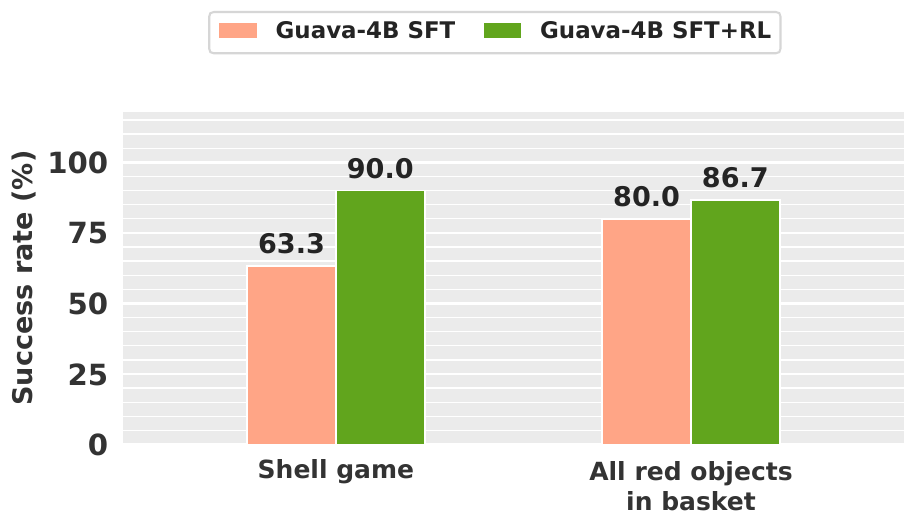}
\captionsetup{font=footnotesize,skip=4pt}
\caption{\textbf{GRPO results.} Success rates over 30 trials per task before and after RL post-training on two long-horizon tasks.}
\label{fig:sft_vs_rl}
\vspace{-8pt}
\end{wrapfigure}

\textbf{Finding 6: Reinforcement learning further improves performance on selected long-horizon tasks.}
We additionally apply GRPO to the SFT checkpoint on two long-horizon tasks, shell game and place all red objects in a basket using settings in Section~\ref{sec:setup}. 
Success rates increase from 63.3\% to 90.0\% and from 80.0\% to 86.7\% respectively (Figure~\ref{fig:sft_vs_rl}). 
These results show that on-policy optimization with environment feedback can further improve the SFT agent on the selected long-horizon tasks.

\par\vspace{4\baselineskip}\WFclear 


\begin{table}[t]
\centering
\small
\caption{\textbf{LIBERO-PRO success rate.} 
We evaluate on the object, goal, and spatial suites under initial-position and task-instruction perturbations, following the LIBERO-PRO evaluation protocol adopted by CaP-X~\cite{fu2026capxframeworkbenchmarkingimproving}.
}
\label{tab:libero}
\setlength{\tabcolsep}{3pt}
\footnotesize
\begin{tabular}{llccccc}
\toprule
Benchmark &
Pert. &
\texttt{Guava-4B} &
\texttt{GPT-5.4} &
\texttt{Qwen3.5-4B} &
$\pi_{0.5}$ &
\texttt{Cap-Agent0}
\\[1pt]
\midrule
\multirow{2}{*}{libero-object} & Pos & 0.60 & 0.61 & 0.13 & 0.17 & 0.22 \\
 & Task & 0.60 & 0.57 & 0.10 & 0.01 & 0.18 \\
\addlinespace
\multirow{2}{*}{libero-goal} & Pos & 0.39 & 0.44 & 0.03 & 0.38 & 0.26 \\
 & Task & 0.38 & 0.47 & 0.00 & 0.00 & 0.17 \\
\addlinespace
\multirow{2}{*}{libero-spatial} & Pos & 0.25 & 0.29 & 0.00 & 0.20 & 0.12 \\
 & Task & 0.26 & 0.26 & 0.00 & 0.01 & 0.14 \\
\midrule
\multirow{2}{*}{Average} & Pos & \underline{0.41} & \textbf{0.45} & 0.06 & 0.25 & 0.20 \\
 & Task & \underline{0.41} & \textbf{0.44} & 0.03 & 0.01 & 0.16 \\
\bottomrule
\end{tabular}
\end{table}

\subsection{Inference Efficiency}
\label{sec:efficiency}

As shown in Table~\ref{tab:efficiency}, local \texttt{Guava-4B} inference on a single RTX 5090 reduces mean model-call latency by a factor of 7.10 and cumulative model-request time per episode by a factor of 7.47 compared to using \texttt{GPT-5.4} API. 
Request counts remain similar, indicating that the reduction in inference waiting time comes primarily from faster responses rather than fewer interaction steps. Generated token volume is also 66.7\% lower. 
Together, these results show that \texttt{Guava-4B} substantially reduces inference overhead and token usage while achieving performance approaching the frontier model.

\begin{table}[t]
\centering
\small
\caption{\textbf{Inference efficiency.}
Means over 30 episodes on three tasks \emph{pick up carrot}, \emph{reverse cube stacking}, \emph{place all red objects in basket} per model, except call latency, which is averaged over requests.}
\label{tab:efficiency}
\setlength{\tabcolsep}{3pt}
\begin{tabular}{@{}lrr@{}}
\toprule
Metric & \texttt{GPT-5.4} & \texttt{Guava-4B} \\
\midrule
Model-call latency (s) $\downarrow$
& 4.168 & \textbf{0.587} \\
Model calls/episode
& 10.10 & 9.60 \\
Model-request time/episode (s) $\downarrow$
& 42.10 & \textbf{5.64} \\
Generated tokens/episode $\downarrow$
& 2389 & \textbf{795} \\
\bottomrule
\end{tabular}
\end{table}

\subsection{Limitations}
\label{sec:limitations}
Agentic manipulation capabilities distilled to \texttt{Guava-4B} remain bounded by the perception and action tools exposed through the shared harness. 
The current toolset supports tabletop manipulation but does not address dexterous in-hand manipulation or high-frequency reactive manipulation.
The single-view observation setup also limits visibility under occlusion and can make spatial relationships ambiguous.
Although local inference reduces mean model-call latency to 0.587\,s, the sequential observation--reasoning--action loop still limits responsiveness to rapidly changing scenes. 
Richer manipulation tools, multi-view observations and latency optimization are promising directions for extending the agent's reliability and task scope.

\section{Related Work}
\label{sec:related}

\textbf{Robotic Manipulation Agents.}
Foundation models support robotic manipulation at different levels of abstraction. 
Vision--language--action (VLA) models directly map visual observations and language instructions to robot actions~\cite{driess2023palme,geminiroboticsteam2025geminiroboticsbringingai,black2025pi,pi0.5}. 
A complementary approach retains the VLM as a high-level decision maker and delegates perception and action execution to reusable modules; learned policies, including VLAs, can themselves serve as such action modules.

Along this direction, earlier systems ground VLM decisions through structured interfaces such as executable programs, affordance or value maps, keypoints, spatial constraints, and learned skills~\cite{liang2022codeap,huang2023voxposer,huang2024rekep,yuan2024robopoint,pan2025omnimanip,li2025hamster}. 
More recent agentic manipulation systems organize these capabilities into interactive harnesses, where frontier VLMs invoke perception and action tools and reason over execution feedback~\cite{shi2025maestro, fu2026capxframeworkbenchmarkingimproving}. 
These systems demonstrate broad manipulation capabilities through reusable tool interfaces. 
We study how such closed-loop tool-use capabilities can be transferred from a frontier VLM to a substantially smaller model while retaining the same manipulation interface.

\textbf{Learning from Teacher-Generated Robot Experience.}
Recent VLMs commonly use supervised fine-tuning (SFT) followed by reinforcement learning (RL) to improve multimodal reasoning~\cite{yang2025r1,team2025kimi}. 
Distillation from stronger foundation models to compact VLMs has also been studied through teacher predictions, intermediate representations, and generated reasoning traces~\cite{xu2024surveyknowledgedistillationlarge,cai2025llava,bousselham2026vold}. 
However, such post-training paradigms remain underexplored for distilling general closed-loop reasoning and tool-use capabilities to robotic manipulation agents.

The general-purpose reasoning capabilities of LLMs and VLMs have motivated their use as data-generation engines for robotics. Several approaches use foundation models to guide or automate real-world robot data collection~\cite{duan2024manipulateanything,ahn2024autort}; however, collecting physical interactions remains costly and difficult to scale. 
Many methods therefore leverage simulation, using foundation models to generate tasks, environments, or demonstrations at scale~\cite{ha2023scaling,wang2024robogen,wang2024gensim,hua2024corlgensim2}. 
At the same time, increasingly large and diverse simulation platforms provide substantially broader environments for robot data generation and evaluation~\cite{robocasa2024,kim2026molmospaces}. 
These VLM-driven data-generation approaches typically use the resulting data to train visuomotor policies from task specifications and observation-action trajectories. 

Our work bridges these two lines of work by using simulation as an interactive environment for post-training a compact VLM-driven manipulation agent.
Simulation allows teacher model decisions to be executed and validated through their physical consequences.
Rather than using only the teacher's observation-action demonstrations, we train a compact tool-using agent from its interleaved reasoning, tool calls, and execution feedback.
The resulting supervision enables the teacher's manipulation capability to be distilled to a compact VLM agent.

\section{Conclusion}
\label{sec:conclusion}

We presented \texttt{Guava}, a framework for distilling tool-using manipulation capabilities from frontier VLMs into compact agents. 
\texttt{Guava} combines a manipulation harness with a simulation-based data generation pipeline that produces executed and validated teacher interactions augmented with perturbation-based recovery trajectories.

Our harness study shows that harness design choices affect manipulation performance across model scales. 
Using GPT-5.4 as the teacher, we train \texttt{Guava-4B} on 2,268 simulation-generated interaction trajectories. 
\texttt{Guava-4B} substantially outperforms its Qwen3.5-4B base model and approaches GPT-5.4 performance across simulation, LIBERO-PRO, and real-world manipulation, with no real-world fine-tuning. 
A controlled ablation shows that perturbation trajectories particularly improve generalization to unseen tasks.
Finally, local \texttt{Guava-4B} inference substantially reduces latency and token usage relative to frontier-model inference.

These results suggest that simulation-generated interaction experience can provide a practical path for transferring frontier-model manipulation capabilities to compact agents for local deployment. 
Extending this approach to broader toolsets, richer observations, and more responsive long-horizon interaction remains an important direction for future work.

\clearpage
\newpage
\bibliographystyle{assets/plainnat}
\bibliography{main}

\clearpage
\startcontents[appendix]

\section*{Appendix Contents}
\printcontents[appendix]{}{1}{}
\newpage
\beginappendix

\appendix

\section{Manipulation Harness Details}
\label{app:tools}

\subsection{Tool Implementation Details}


The following signatures describe the tools available.

\begin{itemize}
\item \textbf{\texttt{get\_position(target\_name)}} returns the estimated object position computed from the segmented point cloud.

\item \textbf{\texttt{get\_position\_and\_size(target\_name)}} returns the estimated object position and dimensions, supporting reasoning about spatial relations and available clearance.

\item \textbf{\texttt{grasp(target\_name)}} estimates a grasp pose , approaches the object, and closes the gripper. This grasp is generated using a PCA-based top-down grasp planner or a learned grasp planner. Tool response returns either \texttt{grasped} when gripper closure is impeded by contact or \texttt{closed} if the gripper closes completely.


\item \textbf{\texttt{align(target\_name, position, clearance)}} positions the gripper relative to an object.
The approach direction is selected from \texttt{top}, \texttt{left}, \texttt{right}, \texttt{front}, and \texttt{back}; the clearance is selected from \texttt{small}, \texttt{medium}, or \texttt{large}, which can be mapped to user-defined values. The current implementation maps the categories to 0.05, 0.08, and 0.15 m.
The tool combines estimated object geometry with clearance to determine a target pose.
Top and side approaches require different geometric clearances.

\item \textbf{\texttt{move(target\_position)}} translates the end effector to an absolute Cartesian position $[x,y,z]$, preserving its orientation.

\item \textbf{\texttt{rotate(angle\_deg, axis)}} rotates the gripper in place by the specified angle in degrees about a gripper-body axis, with \texttt{axis} selected from \texttt{x}, \texttt{y}, and \texttt{z}.

\item \textbf{\texttt{close\_gripper()}} closes the fingers and returns the closure status.

\item \textbf{\texttt{release()}} opens the gripper.

\item \textbf{\texttt{home\_pose()}} returns the robot to its configured home pose, providing a reset action when the current configuration is unsuitable for the next operation.
\end{itemize}

\subsection{Harness Study Details}
\label{app:harness_study}
The harness study in Table~\ref{tab:harness} evaluates seven tasks.
Each model--configuration pair is evaluated over 15 trials per task.
For the multi-action re-planning variant, the model generates a program consisting of multiple tool calls using the same tool set as the full harness, with additional support for variable assignments, \texttt{if} conditionals, and bounded \texttt{for} loops.
The program is executed before the model is queried again.
During execution, intermediate tool returns are available to the program, but intermediate images and automatic tool-response transcripts are not exposed to the model.
Once execution terminates, the model receives a feedback message containing the final image, gripper state, explicitly printed program output, and any runtime error that caused early termination.
If the task is not yet complete, the model may generate another program, for up to five rounds.

\subsection{Harness Evaluation on Additional VLMs}
We further tested several other frontier and open-source models with \texttt{Guava}'s full harness as shown in Table~\ref{tab:sim_baseline_suppl}. 
The two frontier models, \texttt{Gemini-3.1-Pro} and \texttt{Claude-Sonnet-4.6}, achieve high success rates under the same interface, whereas the smaller \texttt{Qwen3.5-2B} performs poorly, with frequent reasoning errors and invalid tool calls observed during evaluation.

\begin{table*}[t!]
    \centering
    \small
    \caption{
    \textbf{Results of Guava harness with additional VLMs.}
    Values are success rates (\%) evaluated over 15 episodes. 
    Best results are marked \textbf{bold}. 
    }
    \label{tab:sim_baseline_suppl}
    \resizebox{\textwidth}{!}{
        \begin{tabular}{l!{\vrule}ccc}
            \thickhline
            \toprule
            & \multicolumn{3}{c}{\textbf{Success Rate (\%)}} \\
            
            \textbf{Task Name}
            & \texttt{Gemini-3.1-Pro}
            & \texttt{Claude-Sonnet-4.6}
            & \texttt{Qwen3.5-2B}~\cite{qwen2026qwen35}
            \\

            \midrule
            \multicolumn{4}{c}{
                \textcolor{gray}{\textit{Seen tasks}}
            } \\
            \rowcolor{gray!11}
            place can in box      & 100.0 & 100.0 & 0.0 \\
            arrange by size       & 100.0 & \textbf{100.0} & 0.0 \\
            \rowcolor{gray!11}
            remove cube from tray & \textbf{100.0} & 100.0 & 0.0 \\
            push basket           & \textbf{66.7} & \textbf{66.7}  & 0.0 \\
            \rowcolor{gray!11}
            close drawer                    & 80.0 & 73.3 & 0.0 \\
            shell game                      & \textbf{93.3} & 86.7 & 0.0 \\
            \rowcolor{gray!11}
            place all red objects in basket & 86.7 & \textbf{100.0} & 0.0 \\

            \midrule
            \multicolumn{4}{c}{
                \textcolor{gray}{\textit{Unseen objects}}
            } \\
            \rowcolor{gray!11}
            pick up carrot      & \textbf{100.0} & \textbf{100.0} & 26.7 \\
            tomato near potato  & \textbf{100.0} & \textbf{100.0} & 0.0 \\
            \rowcolor{gray!11}
            lemon in bin        & \textbf{100.0} & \textbf{100.0} & 0.0 \\
            push pot            & 60.0 & 60.0 & 0.0 \\

            \midrule
            \multicolumn{4}{c}{
                \textcolor{gray}{\textit{Unseen instructions}}
            } \\
            \rowcolor{gray!11}
            stack cube reverse order          & 100.0 & 86.7 & 6.7 \\
            arrange by size reverse direction & 86.7 & 100.0 & 0.0 \\

            \midrule
            \multicolumn{4}{c}{
                \textcolor{gray}{\textit{Unseen long-horizon compositions}}
            } \\
            \rowcolor{gray!11}
            separate food and utensils      & \textbf{100.0} & 86.7 & 0.0 \\
            set table                       & 73.3 & 73.3 & 0.0 \\

            \midrule
            \rowcolor{gray!11}
            \textbf{Overall}
            & \textbf{89.8}
            & 88.9
            & 2.2
            \\

            \bottomrule
            \thickhline
        \end{tabular}
    }
    \vspace{-2.5mm}
\end{table*}

\section{Data Pipeline Details}
\label{app:dataset}

\subsection{Prompt}
\label{app:base_collection}

The data collection system prompt ~\ref{prompt:robot_system} specifies the observation format, available tools, and the reasoning--action response format. Refer to code for full prompts. 

\begin{promptbox}[System Prompt for Data Generation]
\label{prompt:robot_system}
You are an intelligent robot arm controller. You will be shown an image of the scene and given a task to complete using the available tools.

\medskip
\noindent\textbf{Interaction Loop}

In every response, think step-by-step inside \texttt{<think></think>} tags first, then call exactly one tool. When the task is fully complete or irrecoverable, output your thinking and end with \texttt{Task complete} or \texttt{Task failed}.

\medskip
\noindent\textbf{Reasoning Guidelines}

Inside \texttt{<think>}, analyze current scene state - including object and gripper poses, progress of task - what has been done and what remains, result of last action - was it a success, failure, or unexpected outcome, and propose the best next action based on your reasoning...

\medskip
\noindent\textbf{Tool Calling}

\begin{verbatim}
<tool_call>{"name": "tool_name", "arguments": {"param": value}}</tool_call>
\end{verbatim}

\medskip
\noindent\textbf{Gripper State}

The gripper's current position \texttt{[x,y,z]}, rotation \texttt{[roll,pitch,yaw]}, and gripper opening \texttt{\%} are provided at every turn. Use these values directly in your reasoning when necessary.

\medskip
\noindent\textbf{Tool Definition}

...

\end{promptbox}

\subsection{Perturbation Generation Details}
\label{app:branches}

Starting from checkpoints along successful base trajectories, we sample predefined perturbations compatible with the current manipulation stage.
Examples include an object dropped during transport or an object displaced after alignment.
Eligibility depends on the preceding action and physical state: a transport-drop perturbation requires a held object, whereas a grasp-related perturbation applies to a grasp attempt rather than a pushing action.
The teacher observes the perturbed execution outcome and continues interacting until it completes the task or the continuation terminates.

\begin{table}[t]
\centering
\small
\caption{Perturbations used to construct recovery supervision. Perturbations are conditioned on execution stages and will only be applied if compatible.}
\label{tab:appendix_perturbations}
\begin{tabularx}{\linewidth}{@{}lXX@{}}
\toprule
Perturbation & Eligible stage & Information required for recovery \\
\midrule
Missed grasp & Grasp attempt & Whether contact secured the object and how to retry \\
Transport drop & Motion while holding an object & The object's updated location and a new retrieval action \\
Object displacement & After alignment, before the intended grasp & Whether the previous approach remains appropriate \\
Incorrect approach & Object-referenced alignment & A revised approach based on the resulting pose and feedback \\
\bottomrule
\end{tabularx}
\end{table}


We also collect continuations from intermediate task states with a fresh interaction history.
These examples expose the model to partially completed tasks without requiring the original execution prefix.
They do not necessarily involve an execution failure, but instead start from a valid intermediate state along the task trajectory.
All retained rollouts undergo the same outcome verification and quality-control procedure and are counted as perturbation trajectories.

\subsection{Data Cleaning}
\label{app:data_qc}

We provide additional details of the filtering and quality-control procedure summarized in Section~\ref{sec:data_generation_pipeline}.
Automated checks, including rule-based filtering and agent-based inspections, handle most trajectory validation cases.
Rule-based filtering removes malformed or invalid tool arguments and trajectories exceeding max number of tool calls. Agent inspects consistency of reasoning annotations with the instruction, tool call, and execution feedback. 
This includes coordinate-related wording, assumptions about post-release motion, and unsupported statements. 
Reasoning text is corrected when the recorded execution supports the revision; missing reasoning is reviewed against the associated action and available context.
Human review is used when the agent cannot reliably determine the outcome, such as ambiguous object placements or final views in which the relevant objects are partially occluded.

\section{Additional Qualitative Analysis}

\begin{figure}[t]
\centering
\includegraphics[width=0.95\linewidth,trim={0cm 0cm 0cm 0cm},clip]{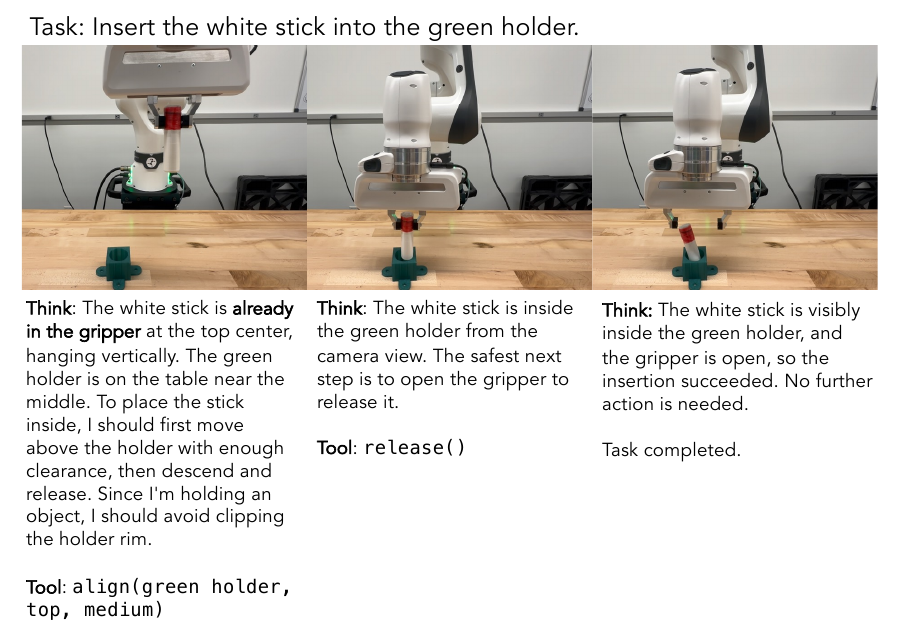}
\caption{
\textbf{Execution-state adaptation.}
When initialized with the peg already grasped, \texttt{Guava-4B} proceeds directly to insertion rather than repeating the completed grasp.
}
\vspace{-2.5mm}
\label{fig:state_aware}
\end{figure}

\paragraph{\textbf{Execution-State Adaptation.}}
In real-world experiments, we observe cases in which the agent resumes an interrupted task without repeating actions that have already been completed. 
For example, if an object is already grasped, the agent proceeds directly to the next step instead of attempting another grasp. 
This behavior suggests that the agent can use the current observation and interaction context to infer task progress. 
As shown in Figure~\ref{fig:state_aware}, when the peg is already grasped, the agent proceeds directly to insertion without issuing another grasp action.

\paragraph{\textbf{Sim2Real Transfer.}} 
\texttt{Guava-4B} transfers to our real-world setup despite being trained entirely from simulation-generated interactions.
One possible contributing factor is that agentic formulation separates high-level semantic planning from low-level perception and control, allowing for effective sim2real transfer. 
This abstraction allows the system to leverage the general visual understanding capability of pretrained VLMs and generalize to new scenes without task-specific robot data.

We also observe individual tasks for which simulation performance is not higher than real-world performance.
Imperfect contact dynamics in simulation may introduce failure modes that do not directly reflect real-world performance.  
This suggests that real-world evaluation remains essential for assessing embodied agentic systems. Future work should further study how to evaluate such systems accurately and efficiently across both simulation and real-world settings.

\begin{figure}[t]
\centering
\includegraphics[width=\linewidth,trim={1.0cm 1.3cm 1.0cm 0.5cm},clip]{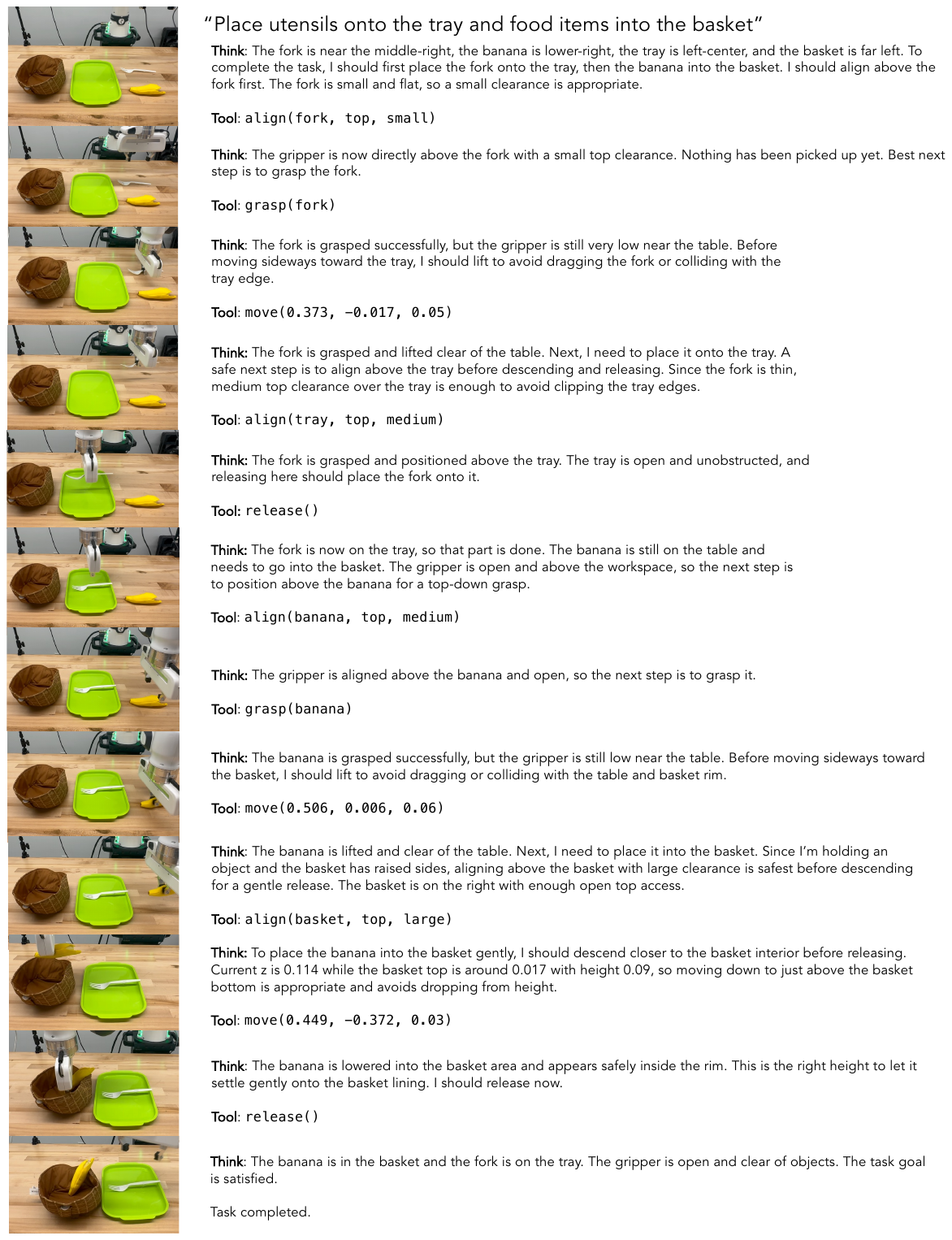}
\caption{
\textbf{Long-horizon execution.}
Example trajectory in which \texttt{Guava-4B} sequentially picks and places multiple objects to complete the task.
}
\label{fig:long_horizon}
\end{figure}

\begin{figure}[t]
\centering
\includegraphics[width=\linewidth,trim={1.0cm 7.5cm 1.0cm 0.5cm},clip]{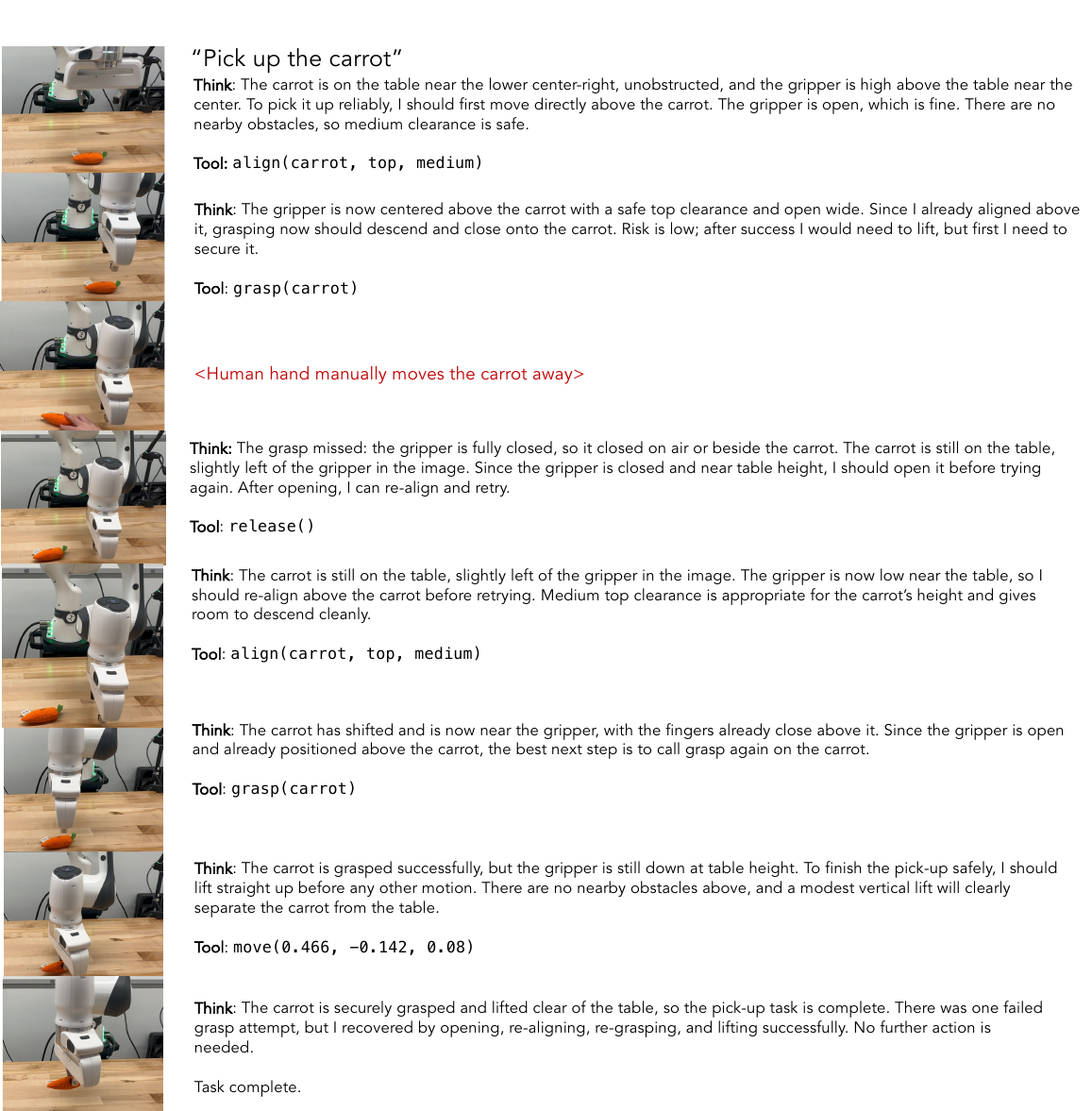}
\caption{
\textbf{Recovery from a failed grasp.}
After an unsuccessful grasp attempt, \texttt{Guava-4B} observes the updated scene, retries the grasp, and continues the task.
}
\label{fig:recovery_example}
\end{figure}

\begin{figure}[t]
\centering
\includegraphics[width=\linewidth,trim={1.0cm 13.5cm 1.0cm 0.5cm},clip]{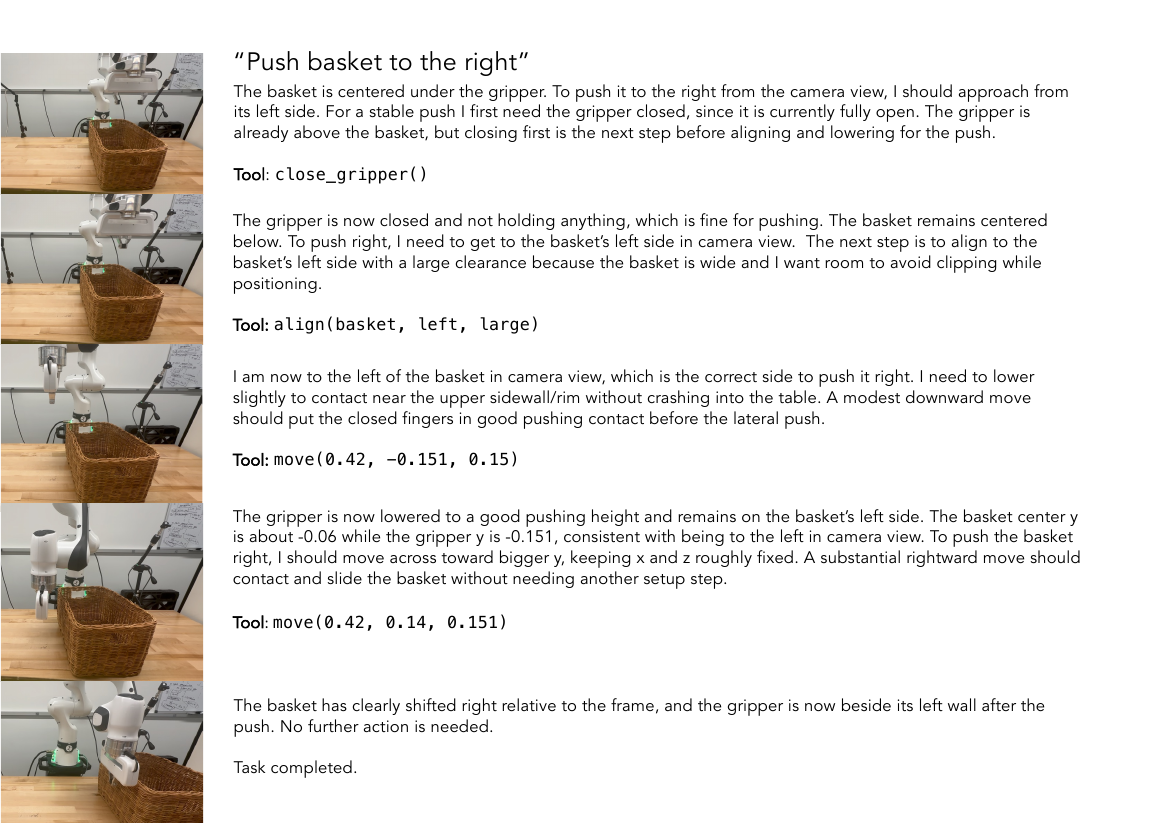}
\caption{
\textbf{Spatially grounded pushing.}
Example trajectory requiring the agent to select an appropriate pre-contact pose and pushing direction relative to the target object.
}
\label{fig:spatial_example}
\end{figure}

\paragraph{\textbf{Representative \texttt{GUAVA-4B} Trajectories.}}
We present example reasoning and tool calling trajectories generated by our method for some tasks in Figure~\ref{fig:long_horizon},\ref{fig:recovery_example} and \ref{fig:spatial_example}. 
These examples illustrate our method’s ability to perform long-horizon planning, recover from execution failures, and reason about spatial relationships in the scene.

\end{document}